\documentclass[journal]{IEEEtran}
\usepackage{graphicx}
\usepackage{float}
\usepackage{subfigure}

\usepackage{amssymb}
\usepackage{amsmath}
\usepackage{url}
\usepackage{multirow}
\usepackage{booktabs}%
\usepackage{threeparttable}

\usepackage{cite}

\ifCLASSINFOpdf
\else
\fi
\begin{document}
%
\title{Real time backbone for semantic segmentation}
%
%
%

\author{Zhengeng~Yang*,
        Hongshan~Yu*,
        Qiang~Fu,
        Wei~Sun,
        Wenyan~Jia,
        Mingui~Sun,
        Zhi-Hong~Mao
\thanks{This work was supported in part by the National Natural Science Foundation of China under Grant 61573135 and U1813205, in part by the National Key Technology Support Program under Grant 2015BAF11B01, in part by the National Key Scientific Instrument and Equipment Development Project of China under Grant 2013YQ140517, in part by China Scholarship Council(201806130030), in part by Key Research and Development Project of Science and Technology Plan of Hunan Province 2018GK2021, in part by the Hunan Key Laboratory of Intelligent Robot Technology in Electronic Manufacturing under Grant 2018001, in part by the Science and Technology Plan Project of Shenzhen City under Grant JCYJ20170306141557198, in part by the Key Project of Science and Technology Plan of Guagdong Province under Grant 2013B011301014, in part by Key Project of Science and Technology Plan of Changsha City kq1801003, and in part by the National Institutes of Health of the United States under Grant R01CA165255.}

\thanks{Z. Yang, H. Yu, Q. Fu, and W. Sun are with the National Engineering Laboratory for Robot Visual Perception and Control Technology, College of Electrical and Information Engineering, Hunan University, Lushan South Rd., Yuelu Dist., 410082, Changsha, China. Z. Yang is also a visiting scholar of University of Pittsburgh. H. Yu is the corresponding author( e-mail: yuhongshancn@hotmail.com).}
\thanks{W. Jia, Z.H. Mao, M. Sun are with the University of Pittsburgh, Pittsburgh, PA, USA. Z.H. Mao is with the department of Electrical and Computer Engineering. W. Jia and M. Sun are with the Department of Neurosurgery.}
\thanks{* These authors contributed equally to this work. H. Yu is the corresponding author( e-mail: yuhongshancn@hotmail.com).}}

%
%

\markboth{Journal of \LaTeX\ Class Files,~Vol.~14, No.~8, Feb~2019}%
{Shell \MakeLowercase{\textit{et al.}}: Bare Demo of IEEEtran.cls for IEEE Journals}
%



\maketitle

\begin{abstract}
The rapid development of autonomous driving in recent years presents lots of challenges for scene understanding. As an essential step towards scene understanding, semantic segmentation thus received lots of attention in past few years. Although deep learning based state-of-the-arts have achieved great success in improving the segmentation accuracy, most of them suffer from an inefficiency problem and can hardly applied to practical applications. In this paper, we  systematically analyze the computation cost of Convolutional Neural Network(CNN) and found that the inefficiency of CNN is mainly caused by its wide structure rather than the deep structure. In addition, the success of pruning based model compression methods proved that there are many redundant channels in CNN. Thus, we designed a very narrow while deep backbone network to improve the efficiency of semantic segmentation. By casting our network to FCN32 segmentation architecture, the basic structure of most segmentation methods, we achieved 60.6\% mIoU on Cityscape val dataset with 54 frame per seconds(FPS) on $1024\times2048$ inputs, which already outperforms one of the earliest real time deep learning based segmentation methods: ENet.
\end{abstract}



\section{Introduction}
\IEEEPARstart{A}{utonomous} driving, a highly expected technology for solving the problem of ever-increasing traffic jam and traffic accidents, has been a dream of human for a long history. To achieve this goal, self-driving vehicles need to understand the scene it placed in accurately, such as locating the road, identifying any traffic signs may existed and so on. Thus, one of the most important tasks towards autonomous driving is to enable the self-driving vehicles to recognize the objects(e.g., road, pedestrians, cars) encountered in the process of driving. This task is very challenging due to the high complexity of driving scenes. However, with the rapid development of deep learning in past few years, multi-class recognition has become much easier to implement with a well-known task called semantic segmentation.
\par
The semantic segmentation refers to the task of assigning each pixel with a semantic label\cite{yu2018methods}. In recent years, the fully convolutional networks based methods have been advancing the frontiers of this field. Up to now, state-of-the-arts(e.g., DPC\cite{chen2018searching}, DeepLabV3\cite{chen2017rethinking}, PSPNet\cite{zhao2017pyramid}) listed in the leader board of Cityscapes dataset\cite{cordts2016cityscapes}, a well-known dataset for street scene understanding, mostly achieve 80\% or more mean Intersection over Union(mIoU). Despite great success achieved, most of these methods focused on improving the accuracy without taking much consideration of efficiency, making these methods unfriendly to practical applications. For example, the PSPNet can only achieve about 1 Frame Per Seconds(FPS) on Titan X GPU card when fed with $1024\times 2048$ high resolution image.
\begin{figure}[ht]
\centering\includegraphics[width=3in]{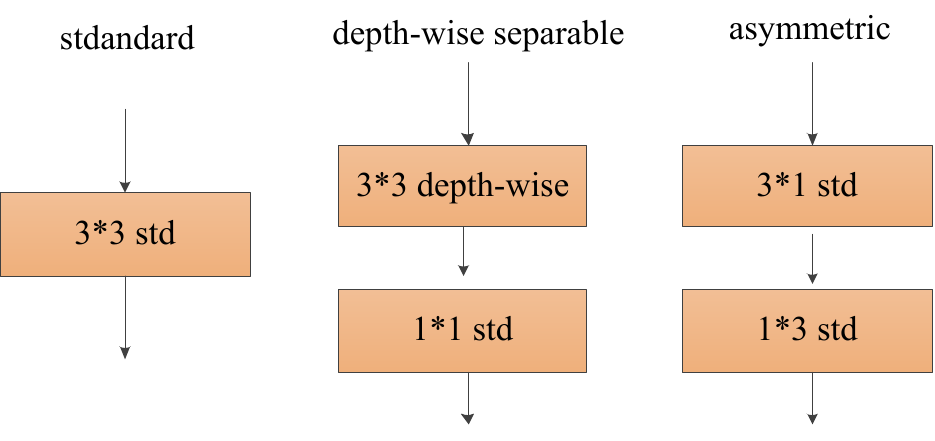}
\caption{Two widely used light weight convolutions. The depth-wise separable convolution factorized a standard $3\times 3$convolution into a depth-wise convolution(detailed in next section) and a $1\times 1$ standard convolution. In contrary, a standard $3\times 3$convolution is factorized into a $3\times 1$ and a $1\times3$ convolution in asymmetric convolution. }
\label{fig_lwc}
\end{figure}
\par
While autonomous driving is highly demanding on efficiency, there has been a growing interest in reducing the computation cost of deep learning based semantic segmentation. A dominate strategy in this direction is to employ Light Weight Convolutions(LWC) to construct deep CNN models, which can reduce the number of parameters significantly while keep the representation ability of deep model almost unchanged. Commonly used LWC including depth-wise separable convolution and asymmetric convolution. For examples, Romera et al.\cite{romera2018erfnet} presented an asymmetric convolution based encoder-decoder segmentation structure, Yu et al\cite{Yu_2018_ECCV}. employed the Xception, a depth-wise separable convolution based network, to capture high-level context for semantic segmentation.

\par
By factorizing a standard $3\times3$ convolution into two separate convolutions(see Fig. \ref{fig_lwc}), the depth-wise separable convolution and asymmetric convolution can reduce the computation cost by a factor of 9(approximately) and 9/6, respectively. However, since the cardinality of computation cost of standard convolution is huge, especially in our case of high resolution street scenes, there are still a lot of rooms for improvement on the efficiency of LWC based networks. To this end, we systematically analyze the computation cost of depth-wise separable convolution and designed a real time backbone network for semantic segmentation. Our contributions are listed as follow
\par
\begin{enumerate}
\item Inspired by the bottleneck structure, we designed a novel real time bottleneck layer by stacking two depth-wise separable convolution and then use the two $1\times 1$ convolution in them to control the number of parameters;
\item We found that the computation cost of CNN largely comes from its width rather than the depth. In addition, the pruning based model compression methods indirectly showed there are redundant widths in deep models. Based on these observations, we designed a very narrow network using our bottleneck layer and achieved 60.6\% mIoU on Cityscape val set with 54.4 FPS.
\end{enumerate}

\par

\par
\section{Narrow Deep Networks}
\label{sec_ndn}
In this section, we first introduce the depth-wise separable convolution, or separable convolution for short, that our narrow deep network based on and analyzed its limitation. Then, we described the motivations behind the designation of the basic unit of our network: narrow bottleneck layer. Finally, we introduce multiple variants of narrow deep networks with different depths and widths.
\subsection{Separable convolution}
\begin{figure}[ht]
\centering\includegraphics[width=2.5in]{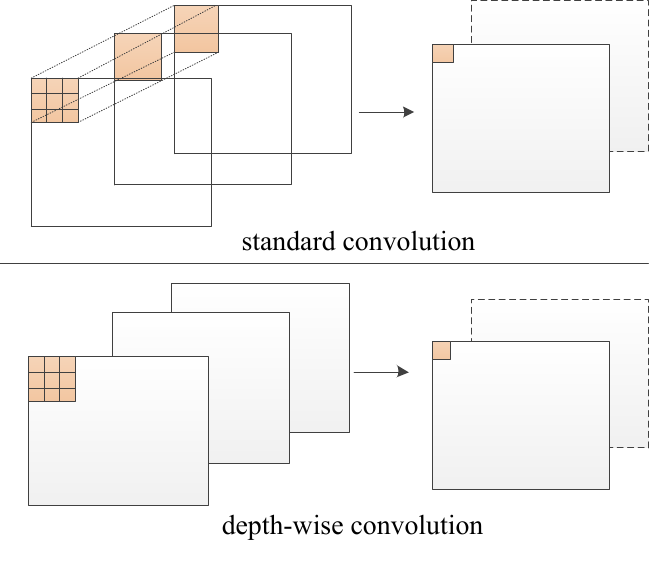}
\caption{Difference between standard convolution and depth-wise convolution. Each output channel computed by standard convolution is correlated with all input channels. In contrast, depth-wise convolution outputs a channel only from a single input.}
\label{fig_sepcov}
\end{figure}

Given a number of feature maps, the standard convolution computes each feature channel from all the input channels. Thus, it has the effect of simultaneously mapping spatial correlations and cross-channel correlations(see Fig.\ref{fig_sepcov}). The separable convolution factorizes a standard convolution into two layers to separately capture spatial correlations and cross-channel correlations. The first layer is called depth-wise convolution and computes feature channel only from a single input channel. The second layer is a $1\times1$ standard convolution(also called point wise convolution), which captures cross-channel correlations by computing linear combinations of the input channels. 

The computation cost of a standard convolution layer in terms of number of ``multiplications+adds(Multi-adds)" is computed by

\begin{equation}
k^2\times H\times W \times M \times N
\label{eq_compcoststd}
\end{equation}

\noindent where $k$ is the kernel size, $H,W$ are the spatial height and width of the feature maps,respectively, $M,N$ are the number of input channel and output channel, respectively. With the same notation, the Multi-adds of separable convolution can be computed by
\begin{equation}
(k^2\times H\times W \times M) +  (H\times W \times M \times N)
\label{eq_compcostsep}
\end{equation}
. The first and second parts split by the plus sign are the computation costs of the depth-wise convolution and $1\times 1$ convolution, respectively. Thus, the computation cost ratio of standard convolution to separable convolution is
\begin{equation}
\frac{k^2N}{k^2+N}
\label{eq_ratio1}
\end{equation}
Since $k$ is usually equal to 3 while $N$ has a very large value(e.g., 512,1024), compared with the standard convolution, the computation cost of separable convolution is reduced approximately by a factor of $k^2=9$. However, there are still a lot of rooms for improvement since the cardinality $HWMN$ is typically in the order of $10^8$. In the following section, we will introduce how to improve the efficiency of separable convolution by incorporating it into residual layers.

\subsection{Narrow Residual layer}

\par
\begin{figure}[ht]
\centering\includegraphics[width=3.3in]{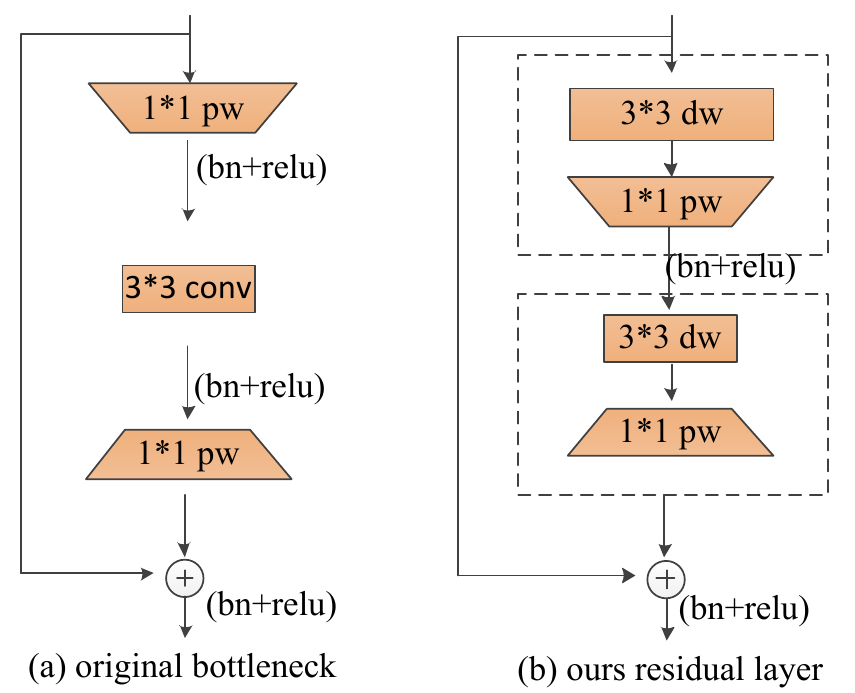}
\caption{Bottleneck residual layers constructed by separable convolution. Based on the idea of original bottleneck, we developed a novel residual layer by stacking two separable convolutions, the two $1\times1$ convolution layer in them are used to achieve the goal of reducing/restoring feature dimension. The widths of the top/bottom edges of the boxes in the Figure represents the width of the input/output channels. The bn and relu represent batch normalization and non-linear activation, respectively. }
\label{fig_residual}
\end{figure}

With the extreme success of ResNet\cite{he2016deep}, the bottleneck residual layer(see Fig. \ref{fig_residual}(a)) has been widely used to control the scale of the model during designing very deep network. The bottleneck layer stacks sequentially a $1\times 1$, $3\times3$, $1\times1$ convolution layer. The two $1\times 1$ layers are responsible for reducing and then restoring feature dimensions. For convenience, we named the three layers in the bottleneck structure as input layer, middle layer and output layer. Denote the input channel numbers of input layer, middle layer, output layer as $n_{in}$, $n_{md}$, $n_{out}$ respectively, we have $n_{in}=n_{out}=n_{md}*e$ in the bottleneck structure, where $e$ is the factor of dimension reduction usually set to 4. Thus, the number of parameters of a three layer bottleneck structure is
\begin{equation}
4n_{md}^2+9n_{md}^2+4n_{md}^2=17n_{md}^2
\label{eq_numberpresidual}.
\end{equation}
In contrast, a standard convolution block having the same width($4n_{md}$) and the same depth($3$) has parameters
\begin{equation}
3\times 9(4n_{md})^2=432n_{md}^2
\label{eq_numberpthreeconv}.
\end{equation}
. Thus, the bottleneck structure is very useful to control the scale of the model if we want to increase the depth of the network.

\par
While the separable convolution contains a $1\times1$ convolution layer in itself, in order to further improve the efficiency of the separable convolution, we stacked two separable convolutions and then used the two $1\times1$ convolutions in them to reduce/restore feature dimensions inspired by the bottleneck structure(see Fig.\ref{fig_residual}(b)). In our case, the middle layer can be viewed as an identity mapping layer.
\par
 Using the same notation as equation (\ref{eq_numberpresidual}), the number of parameters of our residual layer can be computed by
\begin{equation}
N_{layer}=9en_{md}+en_{md}^2+9n_{md}+en_{md}^2
\end{equation}
Thus, the number of parameters of a block contains $d$ residual layers is
\begin{equation}
N=d(2en_{md}^2+(9e+9)n_{md})
\label{eq_numberp_block}
\end{equation}
We can see from the equation that the number of parameters of residual blocks is a quadratic function of $n_{md}$, i.e., the width of the networks, while a linear function of the depth. In addition, according to the computation cost in terms of number of ``Multi+adds" :
\begin{equation}
d(2en_{mid}^2+(9e+9)n_{mid})HW
\label{eq_compcostres}
\end{equation}
where $H,W$ are the size of feature maps, reduction in width will drastically reduce the computation cost since the result of $HW$ are usually higher than ten thousands given high resolution inputs.
\par
Based on analysis above, we argue that it is better to adjust the width rather than the depth for the goal of controlling the scale of model. For example, if we reduce the $n_{md}$ by half, we can have a lighter model even if the depth is doubled. Thus, we proposed to adopt residual layers with very narrow channels to reduce the parameters significantly. The depth of networks is then increased correspondingly to ensure the representative ability of networks. The principle of selection of $n_{md}$ will be detailed in next section.
\begin{figure*}[ht]
\centering\includegraphics[width=5.5in]{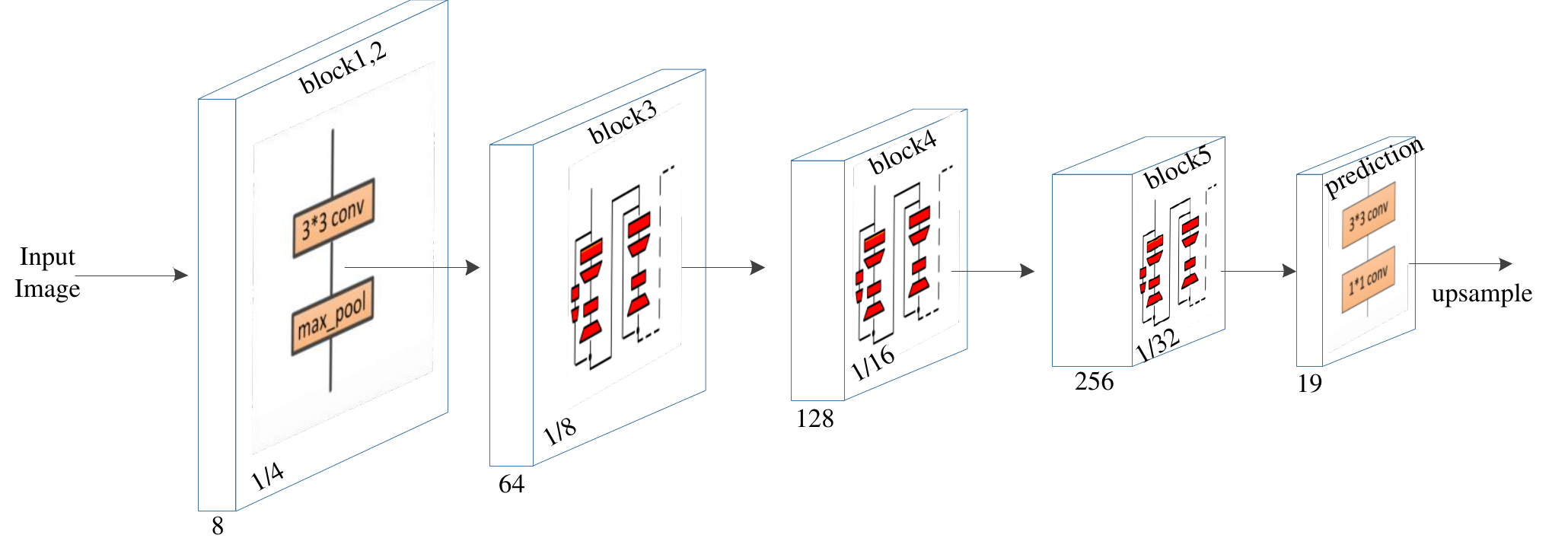}
\caption{FCN32 segmentation structure based on NDNet45. The number below the block represents the number of output channels. The number located at the left corner in the block represents the ratio of feature resolution to the input.}
\label{fig_fcn32}
\end{figure*}

\subsection{Network architecture}
Taking the narrow residual layer as the basic module, we construct three backbone models(showed in Table \ref{table_networktab}) that can be modified to FCN based segmentation models. Similar to most networks, we organized our network into five blocks with different output resolutions. The first two blocks are simply stacked by a $3\times3$ convolution layer and a max pooling layer to reduce the resolution of input quickly. The other three blocks are stacked by residual layers of different numbers $d$. The $res_{n_{md}}$ represents a residual layer that contains $n_{md}$ channels in the middle layer. We fixed the expansion parameter $e$ in all residual layers to 4 so that we can adjust the width of the model by controlling only the $n_{md}$. The number before the $res_{n_{md}}$ is the number of residual layers $d$ in this block. For the convenience of network description, we grouped the $n_{md}$ and $d$ of the last three blocks into $[n_{md1},n_{md2},n_{md3}]$ and $[d1,d2,d3]$, respectively. We named the two groups as channel combination and depth combination.
\par
Next, we introduce how to choose the channel combination and depth combination. Since our goal is to design real-time deep backbone for semantic segmentation, we choose these numbers based on two main rules as follow.
\par
1)The depth of the network is no less than 25. It is well known that the learning capacity of networks is highly related to its width and depth. Since we aimed at reducing the width of networks to improve the efficiency, we restrict our network to a relative large depth to ensure its learning capacity.
\par
2) When casted to FCN32 segmentation structure, the frame processing speed on $1024\times2048$ inputs is no less than 50 Frame per seconds(FPS) on Titan X card. We set this rule because the FCN32 is the most basic structure for deep learning based semantic segmentation. So it is necessary for our network to achieve a very high FPS to leave a large margin for applying other improvement strategies.
\begin{table}[ht]
\renewcommand\arraystretch{1.2}
\caption{Different network architectures constructed by our residual layer }

\centering
\hspace*{-0.2cm}
\begin{tabular}{lllll}
\hline
\toprule
block name  &output scale &  29layer       & 45layer         &61layer  \\
\midrule
conv1      &    1/2      &&3*3 s=2&     \\
max\_pool  &    1/4      &&3*3 s=2&   \\
block1     &    1/8      &$3\times res_{24}$  &$4\times res_{16}$  &$6\times res_{12}$                          \\

block2     &    1/16     &$8\times res_{48}$  &$12\times res_{32}$  &$16\times res_{24}$                       \\

block3     &    1/32     &$3\times res_{96}$  &$6\times res_{64}$  &$8\times res_{48}$                         \\
FPS        &          & 55.23  & 54.34  & 52.36    \\
\bottomrule
\end{tabular}
\label{table_networktab}
\end{table}
\par
Based on the two rules, we first tried a channel combination of $[32,64,128]$ ($n_{md}$), one half of the commonly used $[64,128,256]$. Unfortunately, we found it is difficult for this channel combination to satisfy both of the two rules. So we reduced the channel combination to $[24,48,96]$. Then, we get a 29 layers network by setting the depth combination to $[3,8,3]$ . Note that each separable convolution is viewed as a single layer although it contains a depth-wise convolution and a $1\times1$ convolution. We further choose other two smaller channel combinations: $[12,24,48]$,$[16,32,64]$ to yield deeper backbones(see Table \ref{table_networktab}). To differentiate these networks, we named them as Narrow Deep Network(NDNetX), where $X$ represents the depth.
\par
Based on the NDNet, the FCN32 segmentation architecture can be easily constructed by adding a $1\times 1$ convolution layer to output class scores for each spatial position. However, as proved by DeepLab\cite{chen2014semantic}, it is better to use $3\times 3$ convolution to perform prediction since it considers more local context. Inspired by this insight, we first applied a $3\times 3$ convolution layer over the output of NDNet to aggregate local context and then applied a $1\times 1$ convolution to perform prediction. We showed an example of FCN32 structure constructed by our NDNet45 in Fig .\ref{fig_fcn32}.

\section{Experiments}
\label{sec_exp}
\subsection{Experimental set ups}
\label{sec_expsetup}
\emph{Training policy:}Most semantic segmentation models are trained from a pre-trained classification model such as ResNet. Although such transfer learning strategy can speed up the training, He et al.\cite{he2018rethinking} have proved that it is not necessary to use pre-trained model when adapt the classification network to other tasks. Since our NDNet is first presented, we directly trained our NDNets on semantic segmentation datasets without pre-training them on ImageNet. We trained our models for total 80K steps on the Cityscapes dataset. We set the base learning rate to 0.1 and divided it by 10 at 35K and 60K step respectively.
\par
\emph{Optimizer and loss function:} We trained all our models with the standard stochastic gradient descent(SGD) algorithm with momentum of 0.9. The loss function is defined as the mean of cross entropy terms at each pixel location. Weight decay is also used and the decay parameter is set to 0.9.
\par
\emph{Batch normalization parameters:} Each convolution layer in our network is followed by a Batch Normalization(BN)\cite{ioffe2015batch} layer computed by
 \begin{equation}
\hat{x_i}=\frac{x_i-\mu_i}{\sqrt{Var_i+\epsilon} } \gamma + \beta.
\label{eq_bnormal}
\end{equation}

where $i$ is the index of feature channel, $\gamma$ and $\beta$ are two learnable parameters of linear transform,  $\mu$ and $Var_i$ are the mean and variance of a feature channel, $\epsilon$ is a small constant, Because the concept of batch is invalid during testing, the $\mu_i$ and $Var_i$ are usually replaced by their statistics in this case. The statistics of  $\mu_i$ and $Var_i$ are computed with
\begin{equation}
\mu_{new}=(1-b_{momentum})\times \mu_{old} + b_{momentum} \times \mu_{t}
\end{equation}
during training. Thus, there are two hyper parameters for batch normalization: $\epsilon$ and $b_{momentum}$, we set them to 0.00001,0.1 respectively. In addition, as proved by \cite{wu2018group}, the performance of BN is highly relied on the number of batch size. Thus, we trained our network with batch size 16, image size $1024 \times 1024$.
\par
\emph{Data augmentation:} Following the common rules, we adopted random horizontal mirror and random scales in $[0.75,1,1.25,1.5.2]$ for data augmentation during training.
\par
\emph{Evaluation protocols:}We used mean Intersection of Union (mIoU) and Frame Per Seconds(FPS) to compare the accuracy and efficiency of different models. All the test results reported were obtained from the official evaluation servers, while results on val set were computed by using the codes provided by the official.
\par
\emph{Experiment environments:}
All our experiments were conducted on a computer equipped with a CPU of Intel Xeon E5-1630(8 cores, 3.7 GHz), a GPU of Titan X(12G) and 32G RAM. The construction and training of our deep model were implemented with Pytorch.

\subsection{Cityscapes}
The Cityscapes is a large-scale dataset mainly designed for street scenes understanding. It contains 5,000 finely annotated images and more than 20,000 coarsely annotated images collected from 50 different cities. We considered only the fine annotations for the 19 classes semantic segmentation task. The 5,000 finely annotated images are split into training(2975 images), validation(500 images) and test(1525 images) sets.
%

\begin{table}[ht]
\renewcommand\arraystretch{1.2}
\caption{Comparison of our backbones on Cityscapes validation set}

\centering
\hspace*{-0.2cm}
\begin{tabular}{llllllll}
\hline
\toprule
Network  &max width &mIoU &  params       &FPS                \\
\midrule
wider NDNet29     &1024         &- &3.505M &28.23\\
\midrule
NDNet29           &384         &60.0\%&  515K      &55.23\\
 NDNet29(e=2)     &192        &57.7\% & -&68.30\\
\midrule
NDNet45           &256         &60.6\%&  386K      &54.34\\
NDNet45(e=2)      &128         &57.8\%&  -      &67.02\\
\midrule
NDNet61           &192         &59.5\%&  292K        &52.36\\
NDNet61(e=2)      &96         &56.4\% &    -    &57.47\\
\bottomrule
\end{tabular}
\label{table_compbasic}
\end{table}
\subsection{Comparisons between our backbones}
We first adapted our three backbone networks listed in Table \ref{table_networktab} to FCN32 structure and evaluated their performances. To show the advantages using narrow width, we added a baseline by setting the NDNet29 with a commonly used channel combination([64,128,256]) in last three blocks. However, we cannot train this wider NDNet29 with the batch size mentioned in section\ref{sec_expsetup} limited by GPU memories at hand, so we focused only on the efficiency of the baseline. To also show the influence of width reduction on accuracy, we constructed another group of backbone by setting the fixed expansion parameter to $2$. The experiment results are listed in Table \ref{table_compbasic}. Important conclusions or observations can be drawn as follow
\par
\begin{figure}[ht]
\centering\includegraphics[width=3.3in]{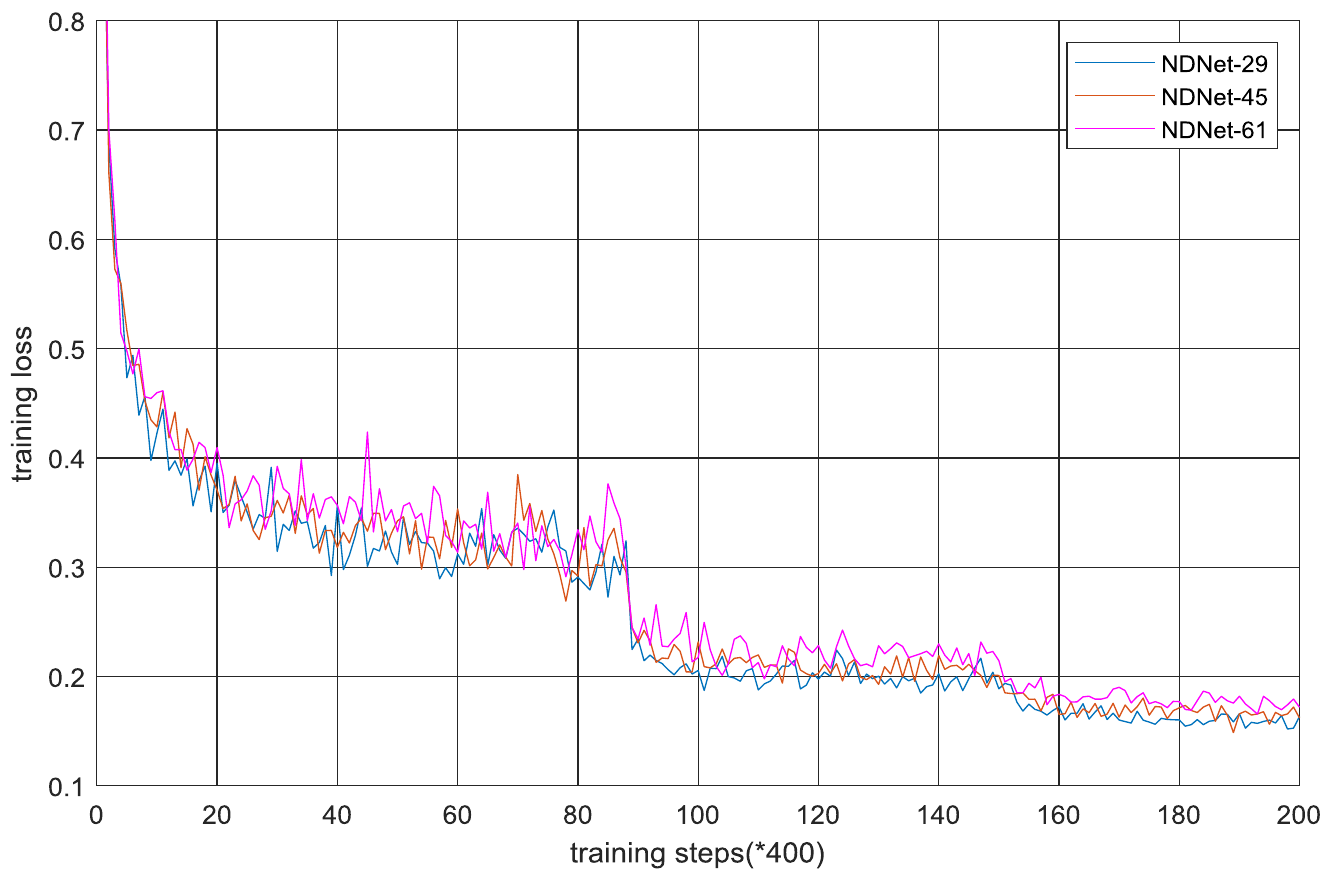}
\caption{Training loss curves of NDNet29(blue), NDNet45(orange) and NDNet61(purple). Although not obviously, the training loss of these three networks generally follows: $NDNet29<NDNet45<NDNet61$. }
\label{fig_loss}
\end{figure}
\par
\begin{enumerate}
\item Our NDNet29 contains about 3.5M parameters when setting with a relative large channel combinations. It is already a light weight model compared with the classical VGG and ResNet that typically contains (40-100)M parameters. This proved the effectiveness of our method on the goal of controlling the scale of network.
\item Wide or narrow: According to the comparison of NDNet29 and its wider version, network with a narrow width outperforms significantly its wider counterpart on the efficiency. Moreover, the narrow version of NDNet29 can achieve 60.0\% mIoU on Cityscape val set, which already outperforms or close to several semantic segmentation method designed for real time goals, such as ESPNetV2(62.7\%).
\item According to the comparison of the three backbones and their narrower version, reducing the width will decrease the accuracy.
\item Width or depth: Different trade-offs between widths and depths show similar performances on both accuracy and efficiency, suggesting that the representative ability of networks after width reduction can be recovered by increasing the depth. In addition, narrower network(NDNet49) can even outperforms wider network(NDNet29) with an appropriate depth. This perhaps because it is easier for wide while shallow network produce over-fitting compared with narrow while deep network(see Fig. \ref{fig_loss}).
\item As mentioned in section \ref{sec_ndn} and proved by this experiment, doubling the depth of the network after halving the width can yield a lighter model. For example, the NDNet29 has 515K parameters while NDNet61 has only 292K parameters. Based on this observation, it is counterintuitive to see that NDNet61 has lower FPS compared with NDNet29. It is reasonable since each convolution layer is followed by a batch normalization operation. In other words, doubling the depth will also doubling the batch normalization operations. Thus, although increasing the depth can recover the learning capacity of a network after width shrink, we cannot increase the depth without restriction.

\end{enumerate}

\par
All these conclusions proved our initial argumentation that it is better to restrict the width rather than the depth when designing network for real time applications. Next, we choose the best performed backbone NDNet45 to compare it with state-of-the-arts and make further improvements.

\section{Conclusions}
\label{sec_conclusion}
Based on the observation that the computation cost of CNN depends largely on the width, we presented a very narrow while deep network for the goal of providing the segmentation community with a real time backbone. Our narrow deep network achieved 60.6\% mIoU on Cityscapes val set with the simplest segmentation structure: FCN32. Most importantly, the FCN32-NDNet achieved 54 FPS on $1024\times 2048$ inputs, leaves a large margin for further improvements. These results show strong evidence that it is better to restrict the width rather than the depth of deep CNN for real time applications.
\par
\bibliography{mybibfile_nodoi}


%
\IEEEpeerreviewmaketitle

\end{document}